\documentclass[letterpaper]{article} % DO NOT CHANGE THIS
\usepackage{aaai25}  % DO NOT CHANGE THIS
\usepackage{times}  % DO NOT CHANGE THIS
\usepackage{helvet}  % DO NOT CHANGE THIS
\usepackage{courier}  % DO NOT CHANGE THIS
\usepackage[hyphens]{url}  % DO NOT CHANGE THIS
\usepackage{graphicx} % DO NOT CHANGE THIS
\usepackage{natbib}  % DO NOT CHANGE THIS AND DO NOT ADD ANY OPTIONS TO IT
\usepackage{caption} % DO NOT CHANGE THIS AND DO NOT ADD ANY OPTIONS TO IT
\usepackage{algorithm}
\usepackage{algorithmic}
\usepackage{amssymb}
\usepackage{amsmath}
\usepackage{booktabs}
\usepackage{multirow}
\usepackage{hyperref}
\hypersetup{pagebackref,colorlinks}
\hypersetup{hidelinks}

\usepackage{newfloat}
\usepackage{listings}
\DeclareCaptionStyle{ruled}{labelfont=normalfont,labelsep=colon,strut=off} % DO NOT CHANGE THIS
\floatstyle{ruled}
\newfloat{listing}{tb}{lst}{}
\floatname{listing}{Listing}
\title{Linear Attention is Enough in Spatial-Temporal Forecasting}
\author{
    Xinyu Ning
}
\affiliations{
    University of Electronic Science and Technology of China\\
    xyning@std.uestc.edu.cn
}

\usepackage{bibentry}
\begin{document}

\maketitle

\begin{abstract}
As the most representative scenario of spatial-temporal forecasting tasks, the traffic forecasting task attracted numerous attention from machine learning community due to its intricate correlation both in space and time dimension. Existing methods often treat road networks over time as spatial-temporal graphs, addressing spatial and temporal representations independently. However, these approaches struggle to capture the dynamic topology of road networks, encounter issues with message passing mechanisms and over-smoothing, and face challenges in learning spatial and temporal relationships separately. To address these limitations, we propose treating nodes in road networks at different time steps as independent spatial-temporal tokens and feeding them into a vanilla Transformer to learn complex spatial-temporal patterns, design \textbf{STformer} achieving SOTA. Given its quadratic complexity, we introduce a variant \textbf{NSTformer} based on Nystr$\ddot{o}$m method to approximate self-attention with linear complexity but even slightly better than former in a few cases astonishingly. Extensive experimental results on traffic datasets demonstrate that the proposed method achieves state-of-the-art performance at an affordable computational cost. Our code is available at \href{https://github.com/XinyuNing/STformer-and-NSTformer}{https://github.com/XinyuNing/STformer-and-NSTformer}.
\end{abstract}


\section{Introduction}

Learning the representation of space and time is long-time vision of machine learning community. Actually, the Convolutional Neural Network (CNN) exploits the spatial information redundancy \cite{he2016deep} while Recurrent Neural Network (RNN) simulates the unidirectionality of time using recurrent structures between neurons.

Traffic forecasting attracted numerous research interest as the most representative scenario of spatial-temporal forecasting, with its intricate correlation both in space and time while broad application \cite{wang2023towards}.

Most works model the traffic road networks to a graph, where the nodes represent the sensors to record traffic conditions such as speed or capacity while the edges represent the topological relationship of nodes, namely roads or distance in most cases. Further, the traffic flow composed of the graph within a period of time can be regarded as a spatial-temporal graph. And the goal of traffic forecasting is to learn a mapping from the past traffic flow to the future.

In the spatial dimension, \cite{li2018diffusion} used CNN to capture the spatial dependencies. Consider the instinct for gird data rather not topology data of CNN, \cite{yu2018spatio} introduced Graph Convolution Network (GCN) \cite{kipf2016semi} to traffic forecasting for learning spatial representation. However, a fixed and static graph is unable to represent the ever-changing road networks, \cite{lin2023dynamic,shao2022decoupled,han2021dynamic} utilize dynamic graph convolution to alleviate the problem. Despite this, the Graph Neural Network (GNN) trends to suffer from over-smoothing problem \cite{chen2020measuring}, while the message passing mechanism between adjacent nodes leads to a deeper network to connect a pair of remote nodes, which cause the parameters of network become harder to be optimized \cite{feng2022powerful}.

In the temporal dimension, \cite{li2018diffusion} and \cite{zhao2017lstm} captured the temporal dependencies using RNN and LSTM respectively.

Thanks to the advantage of parallel processing,capturing long-range dependencies and so on, Transformer \cite{NIPS2017_3f5ee243} has become the de-facto standard not only for natural language processing \cite{devlin2018bert}, but computer vision \cite{dosovitskiy2020image,liu2021swin}, sequential decision \cite{chen2021decision}, and so on.

In the traffic forecasting task, \cite{guo2019attention} proposed an attention-based model. Specifically, it separately used spatial attention and temporal attention first, with GCN in spatial dimension while CNN in temporal dimension behind. And \cite{cai2020traffic} only utilized Transformer to capture the continuity and periodicity of time.

Essentially,all the above works are based on the Spatial-Temporal Graph framework, namely, they modeled the traffic road networks to a graph ( We will continue to use the term in this literature ). We list the inherent drawbacks of the framework here:
\begin{itemize}
    \item First, even using dynamic GNN, it is still hard to capture the spatial dependencies and topological relationship in the complex and ever-changing road networks, needless to say a fixed and static graph.
    \item Second, the GNN trends to suffer from the over-smoothing problem \cite{chen2020measuring}, and the message passing mechanism between adjacent nodes cause that the neural network needs more layers to connect a pair of remote nodes, namely, it increases the difficulty to train the model and optimize the parameters, which become more unbearable in large-scale road networks.
    \item Third, learning the spatial representation and temporal representation separately requires more layers of neural networks to capture the cross-spatial-temporal dependencies.
\end{itemize}

Inspired by the breakthrough of Transformer in graph representation \cite{yun2019graph,chen2022structure,kim2022pure} and time forecasting \cite{zhou2021informer,li2019enhancing,wu2020adversarial}, we study traffic forecasting only using self-attention, namely, we desert any Graph, Convolution and Recurrent module. Obviously, we can immediately overcome the first two out of above problems caused by GNN.

First, we designed a model called \textbf{STformer} (\underline{S}patial-\underline{T}emporal Trans\underline{former}), in which we regard a sensor of road network at a time-step as an independent token rather than a node of graph. We refer to the kind of token as \textbf{ST-Token} because each token is uniquely determined by a time-step and a spatial location. Then the sequence composed by the tokens from the road networks with a period of time is fed to vanilla Transformer. Though the STformer is a extremely concise model, thanks to its ability to capture the cross-spatial-temporal dependencies, it can learn the spatial-temporal representation dynamically and efficiently, and achieves state-of-the-art performance on two most used public datasets \textbf{METR-LA} and \textbf{PEMS-BAY}.

Given the $O(N^2)$ complexity of self-attention, the computational cost of \textbf{STformer} is unbearable under large-scale road networks or long-term forecasting while its performance can be limited. Inspired by \textbf{Nystr{\" o}mformer} \cite{xiong2021nystromformer}, which leverages Nystr{\" o}m method to approximate standard self-attention with $O(N)$ complexity, we designed \textbf{NSTformer} (\underline{N}ystr{\" o}m \underline{S}patial-\underline{T}emporal Trans\underline{former}) with linear complexity. To our surprise, the performance of NSTformer exceeds that of STformer slightly. Actually, this phenomenon gives a open problem to investigate whether approximate attention has other positive effects such as regularization.

We conclude our contributions here:
\begin{itemize}
    \item We investigate the performance of pure self-attention for spatial-temporal forecasting. Our \textbf{STformer} achieves state-of-the-art on METR-LA and PEMS-BAY. We provide a new and extremely concise perspective for spatial-temporal forecasting.
    \item We designed \textbf{NSTformer}, which can achieve state-of-the-art with $O(N)$ complexity. Thanks to the economic linear complexity, we offer the insight that using linear attention to do  spatial-temporal forecasting tasks.
\end{itemize}

\section{Related Work}
\subsection{Transformer in Traffic Forecasting}

We have already discussed the application of deep learning in traffic forecasting task generally in \textbf{Introduction}, particularly the evolution of neural networks for learning the spatial-temporal representation, along with an analysis of the reasons behind it. Here we focus on application of Transformer in traffic forecasting.

\cite{guo2019attention} proposed an attention-based model. Specifically, they designed a ST block, and several blocks are stacked to form a sequence. In each block, spatial attention and temporal attention separately learn the spatial representation and temporal representation in parallel. Subsequently, further learning is performed by GCN and CNN as the same way. \cite{xu2020spatial} alternately set spatial Transformer and temporal Transformer to learn, incorporating GCN in parallel within each spatial Transformer to capture spatial dependencies. \cite{zheng2020gman} combined spatial and temporal attention mechanisms via gated fusion. In summary, these works are still under the Spatial-Temporal Graph framework and learn the spatial representation and temporal representation separately.

\cite{pdformer} did not use any graph structure but designed an intricate Semantic Spatial Attention, Geographic Spatial Attention with Delay-aware Feature Transformation to capture the spatial dependencies while a parallel Temporal self-attention.

\cite{liu2023spatio} is the most related work with ours. They proposed spatial-temporal adaptive embedding to make vanilla Transformer yield outstanding results, rather than designing complex network structures to obtain marginal performance improvements through arduous efforts. But it still learn the spatial representation and temporal representation separately. Although our models are simple yet effective, we have overcome the problem by learning the real spatial-temporal representation simultaneously, which makes our work surpass their performance with a lower computational cost.

\subsection{Efficient Transformer}
Transformer has become the de-facto standard in many applications. However, its core module, self-attention mechanism, has $O(N^2)$ space and time complexity, which limits its performance even feasibility when input is large. The research community has long recognized the problem and numerous works have emerged to speed up the calculation of self-attention \cite{keles2023computational}.

Reformer \cite{kitaev2019reformer}, Big Bird \cite{zaheer2020big}, Linformer \cite{wang2020linformer}, Longformer \cite{beltagy2020longformer}, and routing transformers \cite{roy2021efficient} leveraged a blend of hashing, sparsification, or low-rank approximation to expedite computational processes of the attention scores. Nystr{\" o}mformer \cite{xiong2021nystromformer} and \cite{katharopoulos2020transformers} substituted the softmax-based attention with kernel approximations. Performer \cite{choromanski2020rethinking}, Slim \cite{likhosherstov2021sub}, RFA \cite{peng2021random} used random projections to approximate the computation of attention. SOFT \cite{lu2021soft} and Skyformer \cite{chen2021skyformer} suggested to replace softmax operations with rapidly evaluable Gaussian kernels.

Among them, Nystr{\" o}mformer achieves $O(N)$ complexity. Consider the ample of theoretical groundwork providing analysis and guidance for Nystr{\" o}m method \cite{kumar2009ensemble,gittens2016revisiting,li2010making,kumar2012sampling,si2016computationally,farahat2011novel,frieze2004fast,deshpande2006matrix}, we select Nystr{\" o}mformer as the backbone of NSTformer. Utilizing other sub-quadratic Transformer invariant is an interesting direction for future works.

\section{Problem Setting}

We formally define the traffic forecasting task here.

\textbf{Definition 1} (Road Network). Given the road networks where there is $N$ sensors to capture traffic conditions such as speed. At time-step $t$, then the traffic condition form a tensor ${{X}}_t \in \mathbb{R}^{N \times D}$, where $D$ is feature dimension, generally, $D = 1$ in traffic speed forecasting task.

Note that under Spatial-Temporal Graph framework, the road network is presented by $\mathcal{G}=(\mathcal{V,E},\textbf{\textit{A}})$, where $\mathcal{V}={\{ v_1,...,v_N \}}$ presents the nodes, $\mathcal{E \subseteq V \times V}$ presents the edges, and \textbf{\textit{A}} is the adjacent matrix. In this study, we don't use any graph structure in our models, neither set assumptions about graphs in road network modeling as well.

\textbf{Definition 2} (Traffic Flow). The road network during a period of time $T$ form a traffic flow tensor $\textbf{\textit{X}}=({{X}}_1,{{X}}_2,...,{{X}}_T) \in \mathbb{R}^{T \times N \times D}$.

\textbf{Definition 3} (Traffic Forecasting). As the essence of machine learning is to learn a hypothesis $f$ from a hypothesis class $\mathcal{H}$ \cite{lu2020universal,kidger2020universal,valiant1984theory,livni2014computational}, the deep learning method for traffic forecasting is to learn a mapping from past $T$ time steps' traffic flow to future $T'$ time steps' traffic flow with neural networks as follow:

\begin{center}
    \begin{equation}
        [{{X}}_{t-T+1},...,{{X}}_t] \stackrel{f}{\longrightarrow} [{{X}}_{t+1},...,{{X}}_{t+T'}]
    \end{equation}
\end{center}

Correspondingly, the learning of Spatial-Temporal Graph framework is as follow:

\begin{center}
    \begin{equation}
        [{{X}}_{t-T+1},...,{{X}}_t;\mathcal{G}] \stackrel{f}{\longrightarrow} [{{X}}_{t+1},...,{{X}}_{t+T'}]
    \end{equation}
\end{center}

\section{Architecture}

\begin{figure*}[htbp]
    \centering
    \includegraphics[scale=0.45]{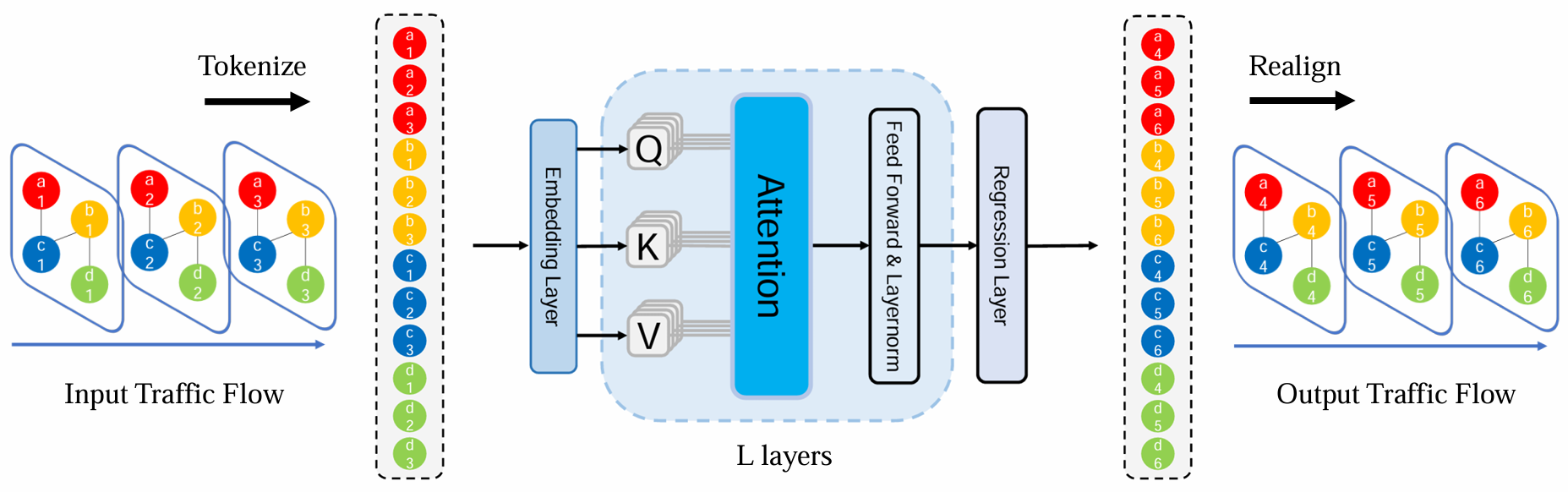}
    \caption{The Architecture of STformer and NSTformer}
    \label{fig:arch}
\end{figure*}

\subsection{Pipeline}

We present our pipeline and the architecture of our models in \textbf{Figure 1}. Without any complicated module or data process, we focus on how to capture the complex spatial-temporal relationships. Specifically, instead of regarding the traffic flow as a spatial-temporal graph, we just treat it as a regular 3D tensor and flatten it to a 1D sequence then feed to Transformer or its variant.

By this way, we can effectively capture the relationship between any pair of ST-Tokens. Correspondingly, the complexity of STformer is $O(N^2T^2)$, which is unbearable when the input is too large. To overcome it, we desinged NSTformer with $O(NT)$ complexity, yielding a powerful and efficient model. And the only one difference between NSTformer and STformer is the attention mechanism, where the former with the linear Nystr{\" o}m attention while the latter with the quadratic self-attention.

\subsection{Embedding Layer \& Regression Layer}
As the \textbf{Figure 1} shows, our models are so concise that it just contains the embedding layers, attention mechanism and regression layers. And the only difference between STformer and NSTformer is their attention mechanism. We first introduce their common module, embedding layers and regression layers and present the models in detail later.

We follow \cite{liu2023spatio} to set our embedding layers, in which they proposed a spatial-temporal adaptive embedding $E_a$ to capture the intricate spatial-temporal dependency rather than using any graph embedding.

Given the traffic flow $\textit{\textbf{X}} \in \mathbb{R}^{T \times N \times D}$, where $T$ is the length of the input time-steps and $N$ represents the number of sensors of road networks, the mainstream setting of the feature dimension $D$ is 3 which contains the value of traffic conditions such as speed, the time flag day-of-week from 1 to 7, the time flag timestamps-of-day from 1 to 288. We embed the three features to $E_f \in {\mathbb{R}^{T \times N \times 3d_f}}$, where $d_f$ is the dimension of the feature embedding. And the simple yet effective $E_a \in \mathbb{R}^{T \times N \times d_a}$ to capture intricate spatial-temporal dependencies. After the above embedding, the traffic flow then be $\textit{\textbf{X}} \in \mathbb{R}^{T \times N \times d_h}$, where $d_h$ equals to $3d_f + d_a$, we will introduce our specific setting of the embedding dimensions in \textbf{Experiment}.

After the attention module in STformer and NSTformer, the traffic flow will be mapped to $\textit{\textbf{X'}} \in \mathbb{R}^{T \times N \times d_h}$. Finally, our fully-connected Regression Layer $FC$ yields the prediction $\textbf{\textit{Y}} = FC(\textbf{\textit{X'}}) \in \mathbb{R}^{T' \times N \times d}$, where $T'$ is the length prediction and the $d$ is the dimension of prediction value, which equals to 1 in our traffic forecasting setting.

\subsection{STformer}
In this paper, we refer to the standard self-attention in Transformer as \textbf{self-attention}, the approximated self-attention in NSTformer as \textbf{Nystr{\" o}m attention}.

As \textbf{Figure 1} shows, we regard each nodes (sensors) at different time-steps as an independent \textbf{ST-Token}, and all ST-Tokens from one traffic flow form a sequence whose length is $N \times T$ to feed to attention. Here we present the traffic flow as $\textit{\textbf{X}} \in \mathbb{R}^{NT \times d_h}$. We have 
\begin{center}
\begin{equation}   Q=\textbf{\textit{X}}W_Q, K=\textbf{\textit{X}}W_K, V=\textbf{\textit{X}}W_V\label{1}
\end{equation}   
\end{center}
where $W_Q, W_K, W_V \in \mathbb{R}^{d_h \times d_h}$.

Then the score of the self-attention is calculated as:
\begin{center}
\begin{equation}
    S=softmax(\frac{QK^{T}}{\sqrt{d_h}}) \in \mathbb{R}^{NT \times NT}\label{(4)}
\end{equation}
\end{center}

The final output of the self-attention is:
\begin{center}
\begin{equation}
    SelfAttention(Q,K,V) = softmax(\frac{QK^{T}}{\sqrt{d_h}}) V \in \mathbb{R}^{NT \times d_h}\label{2}
\end{equation}
\end{center}

The \textbf{equation \eqref{(4)}} reveals that STformer can capture the spatial-temporal dependencies simultaneously and dynamically learn the spatial-relationship of the global space, which overcome the problems of Spatial-Temporal Graph frame work.

In the other hand, due to the quadratic complexity in equal (4), the computational complexity of STformer is $O(N^2T^2)$, which limits the performance even feasibility of STformer when the input is large.

\subsection{NSTformer}

To overcome the above new obstacle, we designed NSTformer, in which we adapted Nystr{\" o}mformer to replace the standard Transformer in STformer, yields a linear Nystr{\" o}m attention. Consider the only one difference of the two models is the attention mechanism, we generally introduce Nystr{\" o}mformer \cite{xiong2021nystromformer} and analysis the linear Nystr{\" o}m attention here, we recommend \cite{xiong2021nystromformer} to learn more.

The Nystr{\" o}m-like methods approximate a matrix by sampling columns from the matrix. \cite{xiong2021nystromformer} adapted the method to approximate the calculation of the original softmax matrix $S$ in equal (4). The fundamental insight involves utilizing landmarks $\tilde{K}$ and $\tilde{Q}$ derived from key $K$ and query $Q$ to formulate an efficient Nystr{\" o}m approximation without accessing the entire $QK^T$. In cases where the count of landmarks, $m$, much smaller than the sequence length $n$, the Nystr{\" o}m approximation exhibits linear scalability concerning both memory and time with respect to the input sequence length.

\textbf{Definition 4} \cite{xiong2021nystromformer}. Assume that the selected landmarks for inputs $Q=[q_1;...;q_n]$ and $K=[k_1;...;k_n]$ are $\{ \tilde{q_j} \}_{j=1}^m$ and $\{ \tilde{k_j} \}_{j=1}^m$ respectively. We denote the matrix form of the corresponding landmarks as
\begin{center}
For $\{ \tilde{q_j} \}_{j=1}^m, \tilde{Q}=[\tilde{q_1};...;\tilde{q_m}] \in \mathbb{R}^{m \times d_h}$

For $\{ \tilde{k_j} \}_{j=1}^m, \tilde{K}=[\tilde{k_1};...;\tilde{k_m}] \in \mathbb{R}^{m \times d_h}$
\end{center}

Then the $ m \times m $ matrix is given by $A_S$ = softmax$(\frac{\tilde{Q}\tilde{K}^{T}}{\sqrt{d_h}})$.

And the Nystr{\" o}m form of the softmax matrix, $S=softmax(\frac{QK^{T}}{\sqrt{d_h}})$ is approximated as
\begin{center}
    $\hat{S}$ = softmax$(\frac{Q\tilde{K}^{T}}{\sqrt{d_h}}) {A_S}^+ $softmax$(\frac{\tilde{Q}K^{T}}{\sqrt{d_h}})$,
\end{center}
where ${A_S}^+$ is a Moore-Penrose pseudoinverse of $A_S$.

\textbf{Lemma 1} \cite{xiong2021nystromformer}. For $A_S \in \mathbb{R}^{m \times m}$, the sequence $\{ {Z_j} \}_{j=0}^{j=\infty}$ generated by \cite{razavi2014new},
\begin{center}
\begin{equation}
    Z_{j+1} = \frac{1}{4}Z_{j}(13I-A_SZ_j(15I-A_SZ_j(7I-A_SZ_j)))
\end{equation}
\end{center}
converges to the Moore-Penrose inverse ${A_S}^+$ in the third order with initial approximation $Z_0$ satisfying $||A_S{A_S}^+-A_SZ_0|| < 1$.

By $Z^*$ with (6) in \textbf{Lemma 1} to approximate ${A_S}^+$, then the Nystr{\" o}m approximation of $S$ becomes
\begin{center}
    $\hat{S}$ = softmax$(\frac{Q\tilde{K}^{T}}{\sqrt{d_h}}) Z^* $softmax$(\frac{\tilde{Q}K^{T}}{\sqrt{d_h}})$.
\end{center}

Finally, we derive the Nystr{\" o}m attention:
\begin{center}
    $\hat{S}V$ = softmax$(\frac{Q\tilde{K}^{T}}{\sqrt{d_h}}) Z^* $softmax$(\frac{\tilde{Q}K^{T}}{\sqrt{d_h}})V$.
\end{center}

We present the pipeline for Nystr{\" o}m approximation of softmax matrix in self-attention in \textbf{Algorithm 2}.

\subsubsection{Landmarks selection}

\cite{10.1145/1390156.1390311,NEURIPS2020_f6a8dd1c} used K-Means to select landmark points \cite{pmlr-v97-lee19d}. Consider the EM-style updates in K-means is less preferable when using mini-batch training. \cite{xiong2021nystromformer} suggested using Segment-means, which is similar to the local average pooling approach previously utilized in NLP literature \cite{shen-etal-2018-baseline}.

Particularly, for inputs $Q=[q_1;...;q_n]$ and $K=[k_1;...;k_n]$, $n$ queries are divided into $m$ segments. Assuming $n$ is divisible by $m$ for simplicity, as we can pad inputs to a length divisible by $m$, let $l = n / m$. Landmark points for $Q$ and $K$ are then calculated as demonstrated in \eqref{(7)}. And the whole process only need a simple scan of inputs, which yields a complexity of $O(n)$.

\begin{equation}
    \tilde{q_j} = \sum_{i=(j-1) \times l +1}^{(j-1) \times l +m} \frac{q_i}{m},  \tilde{k_j} = \sum_{i=(j-1) \times l +1}^{(j-1) \times l +m} \frac{k_i}{m} \label{(7)}
\end{equation}

We revisit the landmarks selection for spatial-temporal forecasting. Our insight is that at the same time, the nodes from the same one neighborhood of the road network have similar traffic conditions. 

The hypothesis is reasonable, as the traffic conditions in a localized area are influenced by similar factors such as road capacity, traffic signals or nearby events. For example, in the downtown area of a city, which typically serves as a hub for business and commercial activities, the traffic patterns during rush hours might be characterized by high congestion due to the influx of commuters. This congestion is likely to spread across multiple nodes within the same vicinity, affecting adjacent streets and intersections. Conversely, in residential neighborhoods, the traffic state may be different, with peak times coinciding with school drop-offs and pickups or evening commutes, but generally experiencing lighter traffic compared to commercial districts.

Based on the understanding of spatial-temporal data, we introduce a landmarks selection algorithm for using Nystr{\" o}m attention in spatial-temporal forecasting tasks, which named after \textbf{Spatial-Temporal Cluster Sampling (STCS)} algorithm. Specifically, the whole nodes of road network are divided into clusters $c_1$ to $c_s$ according their distance, and during the $T$ time-steps, each cluster is regarded as a block which we term it \textbf{ST-block}. One can select suitable clustering algorithm to the datasets as long as it can cluster the nodes according their spatial relationships, in our experiment we use the Agglomerative Clustering from scikit-learn \cite{sklearn.AgglomerativeClustering}. As the above hypothesis, we assume the traffic condition of nodes from the same block follows a same Normal distribution. Then, one can sample a certain number of times to get the average value. Finally, $s \times T$ landmarks are selected, as presented in \textbf{Algorithm 1} taking the example of computation of $\tilde{Q}$, that of $\tilde{K}$ has the same process. In this way, we add our insight for spatial-temporal data to landmarks selection. Note that the clustering operation can be finished before the landmarks selection once the road network is given, then we can immediately figure out that the procession of STCS also only require a simple scan for inputs as Segment-means, with $O(n)$ complexity.

\begin{algorithm}
\caption{Spatial-Temporal Cluster Sampling}
\label{alg:algorithm1}
\textbf{Input}: Query $Q$\\
\textbf{Parameter}: Time steps $T$, sampling iterations $p$, Clusters $c_1, c_2, ..., c_s$.\\
\textbf{Output}: $\tilde{Q}$
\begin{algorithmic}[1] %[1] enables line numbers

\STATE Let $\tilde{Q} = [ \ \ ]$.\\
\FOR {$t \gets 1$ to $T$}
    \FOR{$c \gets c_1$ to $c_s$}
        \STATE Compute the mean $\mu$ and the standard variation $\sigma$ of the cluster $c$.\\
        \STATE Let $Sum = 0$.\\
        \FOR{$i \gets 1$ to $p$}
            \STATE Draw one sample $X \sim \mathcal{N}(\mu,\,\sigma^{2})$.\\
            \STATE $Sum = Sum + X$.
        \ENDFOR
        \STATE $Sum = \frac{Sum}{p}$.
        \STATE Append $Sum$ to $\tilde{Q}$.
    \ENDFOR
\ENDFOR
\STATE \textbf{return} $\tilde{Q}=[\tilde{q_1};...;\tilde{q_m}]$, where $m = s \times T$.
\end{algorithmic}
\end{algorithm}

\subsubsection{Complexity analysis}
We follow \cite{xiong2021nystromformer} to analyze the complexity of Nystr{\" o}m attention, namely $\hat{S}V$.

The time complexity breakdown is as follows:

\begin{itemize}
    \item Landmark selection using Segment-means takes $O(n)$.
    \item Iterative approximation of the pseudoinverse takes $O(m^3)$ in the worse case.
    \item Matrix multiplication softmax$(\frac{Q\tilde{K}^{T}}{\sqrt{d_h}})  \times Z^*$, softmax$(\frac{\tilde{Q}K^{T}}{\sqrt{d_h}}) \times V$, (softmax$(\frac{Q\tilde{K}^{T}}{\sqrt{d_h}})  \times Z^*$) $\times$ (softmax$(\frac{\tilde{Q}K^{T}}{\sqrt{d_h}}) \times V$) take $O(nm^2 + mnd_h + m^3 + nmd_h)$.
\end{itemize}

Then we have the overall time complexity $O(n + m^3 + nm^2 + mnd_h + m^3 + nmd_h)$.

The memory complexity breakdown is as follows:

\begin{itemize}
    \item Storing landmarks matrix $\tilde{Q}$ and $\tilde{K}$ takes $O(md_h)$.
    \item Storing four Nystr{\" o}m approximation matrices takes $O(nm + m^2 + mn + nd_h)$.
\end{itemize}

Then we have the overall memory complexity $O(md_h + nm + m^2 + mn + nd_h)$.

Obviously, when $m \ll n$, the time and memory complexity of Nystr{\" o}m attention are $O(n)$.

Correspondingly, the computational complexity of NSTformer is $O(NT)$.
\renewcommand{\algorithmicrequire}{\textbf{Input:}}
\renewcommand{\algorithmicensure}{\textbf{Output:}}

\begin{algorithm}
\caption{Pipeline for Nystr{\" o}m approximation of softmax matrix in self-attention \cite{xiong2021nystromformer}}
\label{alg2}
\textbf{Input}: Query $Q$ and Key $K$.\\
\textbf{Output}: Nystr{\" o}m approximation of softmax matrix.
\begin{algorithmic}[1]
\STATE Compute landmarks from input $Q$ and landmarks
from input $K$, $\tilde{Q}$ and $\tilde{K}$ as the matrix form.

\STATE Compute $\tilde{F}=$ softmax$(\frac{Q\tilde{K}^{T}}{\sqrt{d_h}})$, $\tilde{B}=$ softmax$(\frac{\tilde{Q}K^{T}}{\sqrt{d_h}})$.

\STATE Compute $\tilde{A}=$ softmax$(\frac{\tilde{Q}\tilde{K}^{T}}{\sqrt{d_h}})^+$.

\STATE \textbf{return} $\tilde{F} \times \tilde{A} \times \tilde{B}$.
\end{algorithmic}
\end{algorithm}

\section{Experiment}

\begin{table*}[htpb]
	\renewcommand\arraystretch{0.98}
		\centering
		\setlength{\abovecaptionskip}{0.cm}
		\caption{Experimental Results on METR-LA and PEMS-BAY.}
		\label{tab:main}
		\begin{tabular}{cclll|lll|lll}
		  \toprule
		  \midrule
		  \multirow{2}*{\textbf{Datasets}} &\multirow{2}*{\textbf{Models}} & \multicolumn{3}{c}{\textbf{Horizon 3 (15 mins)}} & \multicolumn{3}{c}{\textbf{Horizon 6 (30 mins)}}& \multicolumn{3}{c}{\textbf{Horizon 12 (60 mins)}}\\ 
		  \cmidrule(r){3-5} \cmidrule(r){6-8} \cmidrule(r){9-11}
		  &  & MAE & RMSE & MAPE & MAE & RMSE & MAPE & MAE & RMSE & MAPE\\
		  \midrule
		  \midrule
		  \multirow{21}*{\textbf{METR-LA}} 
            &HA              & 4.79  & 10.00 & 11.70\%       & 5.47  & 11.45 & 13.50\%      & 6.99  & 13.89 & 17.54\% \\
            &HI              & 6.80  & 14.21 &16.72\%        & 6.80  & 14.21 & 16.72\%      & 6.80  & 14.20 & 10.15\% \\		 
		  &VAR             & 4.42  & 7.80  & 13.00\%       & 5.41  & 9.13  & 12.70\%      & 6.52  & 10.11 & 15.80\% \\ 
		  &SVR             & 3.39  & 8.45  & 9.30\%        & 5.05  & 10.87 & 12.10\%      & 6.72  & 13.76 & 16.70\% \\ 
		  &FC-LSTM         & 3.44  & 6.30  & 9.60\%        & 3.77  & 7.23  & 10.09\%      & 4.37  & 8.69  & 14.00\% \\ 
		  &DCRNN           & 2.77  & 5.38  & 7.30\%        & 3.15  & 6.45  & 8.80\%       & 3.60  & 7.60  & 10.50\% \\ 
            &AGCRN           & 2.85  & 5.53  & 7.63\%        & 3.20  & 6.52  & 9.00\%       & 3.59  & 7.45  & 10.47\% \\
		  &STGCN           & 2.88  & 5.74  & 7.62\%        & 3.47  & 7.24  & 9.57\%       & 4.59  & 9.40  & 12.70\% \\ 
            &STSGCN          & 3.31  & 7.62  & 8.06\%        & 4.13  & 9.77  & 10.29\%      & 5.06  & 11.66 & 12.91\% \\
		  &GWNet           & 2.69  & 5.15  & 6.90\%        & 3.07  & 6.22  & 8.37\%       & 3.53  & 7.37  & 10.01\% \\
            &MTGNN           & 2.69  & 5.18  & 6.88\%        & 3.05  & 6.17  & 8.19\%       & 3.49  & 7.23  & 9.87\% \\  
		  &GTS             & 2.75  & 5.27  & 7.21\%        & 3.14  & 6.33  & 8.62\%       & 3.59  & 7.44  & 10.25\% \\  
		  &ASTGCN          & 4.86  & 9.27  & 9.21\%        & 5.43  & 10.61 & 10.13\%      & 6.51  & 12.52 & 11.64\% \\         
		  &GMAN            & 2.80  & 5.55  & 7.41\%        & 3.12  & 6.49  & 8.73\%       & 3.44  & 7.35  & 10.07\% \\  	  
            &STID            & 2.82  & 5.53  & 7.75\%        & 3.19  & 6.57  & 9.39\%       & 3.55  & 7.55  & 10.95\% \\
            &STNorm          & 2.81  & 5.57  & 7.40\%        & 3.18  & 6.59  & 8.47\%       & 3.57  & 7.51  & 10.24\% \\
            &PDFormer        & 2.83  & 5.45  & 7.77\%        & 3.20  & 6.46  & 9.19\%       & 3.62  & 7.47  & 10.91\% \\
            &STAEformer      & 2.65  & 5.11$^*$  & 6.85\%        & 2.97$^*$  & 6.00$^*$  & 8.13\%       & 3.34  & 7.02$^*$  & 9.70\% \\
            &ST-Mamba      & 2.64$^*$  & 5.17  & \textbf{6.61}\%$^*$        & 3.00  & 6.18  & 8.06\%$^*$       & 3.33$^*$  & 7.10  & 9.56\%$^*$ \\
            \cmidrule(r){2-11}
            &\textbf{STformer}  & \underline{2.60} & \underline{4.90} & 6.75\% & \underline{2.84}  & \underline{5.59}  & \underline{7.75}\%  & \underline{3.12} & \underline{6.45}  & \underline{8.93}\% \\
            \cmidrule(r){2-11}
            &\textbf{NSTformer} &  \textbf{2.58} & \textbf{4.87} & 6.69\%      & \textbf{2.82}  & \textbf{5.55}  & \textbf{7.64}\%       & \textbf{3.09}  & \textbf{6.39}  & \textbf{8.78}\% \\
		\midrule
		\midrule

		\multirow{21}*{\textbf{PEMS-BAY}} 
            &HA              & 1.89  & 4.30  & 4.16\%        & 2.50  & 5.82  & 5.62\%       & 3.31  & 7.54  & 7.65\% \\ 
            &HI              & 3.06  & 7.05  & 6.85\%        & 3.06  & 7.04  & 6.84\%       & 3.05  & 7.03  & 6.83\% \\		  
		  &VAR             & 1.74  & 3.16  & 3.60\%        & 2.32  & 4.25  & 5.00\%       & 2.93  & 5.44  & 6.50\% \\ 
		  &SVR             & 1.85  & 3.59  & 3.80\%        & 2.48  & 5.18  & 5.50\%       & 3.28  & 7.08  & 8.00\% \\ 
		  &FC-LSTM         & 2.05  & 4.19  & 4.80\%        & 2.20  & 4.55  & 5.20\%       & 2.37  & 4.96  & 5.70\% \\ 
		  &DCRNN           & 1.38  & 2.95  & 2.90\%        & 1.74  & 3.97  & 3.90\%       & 2.07  & 4.74  & 4.90\% \\ 
            &AGCRN           & 1.35  & 2.88  & 2.91\%        & 1.67  & 3.82  & 3.81\%       & 1.94  & 4.50  & 4.55\% \\
		  &STGCN           & 1.36  & 2.96  & 2.90\%        & 1.81  & 4.27  & 4.17\%       & 2.49  & 5.69  & 5.79\% \\ 
            &STSGCN          & 1.44  & 3.01  & 3.04\%        & 1.83  & 4.18  & 4.17\%       & 2.26  & 5.21  & 5.40\% \\ 
		  &GWNet           & 1.30  & 2.74  & 2.73\%        & 1.63  & 3.70  & 3.67\%       & 1.95  & 4.52  & 4.63\% \\
            &MTGNN           & 1.32  & 2.79  & 2.77\%        & 1.65  & 3.74  & 3.69\%       & 1.94  & 4.49  & 4.53\% \\  
		  &GTS             & 1.34  & 2.83  & 2.82\%        & 1.66  & 3.78  & 3.77\%       & 1.95  & 4.43  & 4.58\% \\  
		  &ASTGCN          & 1.52  & 3.13  & 3.22\%        & 2.01  & 4.27  & 4.48\%       & 2.61  & 5.42  & 6.00\% \\        
		  &GMAN            & 1.34  & 2.91  & 2.86\%        & 1.63  & 3.76  & 3.68\%       & 1.86$^*$  & 4.32$^*$  & 4.37\%$^*$ \\  		  
            &STID            & 1.31  & 2.79  & 2.78\%        & 1.64  & 3.73  & 3.73\%       & 1.91  & 4.42  & 4.55\% \\
            &STNorm          & 1.33  & 2.82  & 2.76\%$^*$        & 1.65  & 3.77  & 3.66\%       & 1.92  & 4.45  & 4.46\% \\
            &PDFormer        & 1.32  & 2.83  & 2.78\%        & 1.64  & 3.79  & 3.71\%       & 1.91  & 4.43  & 4.51\% \\
            &STAEformer      & 1.31  & 2.78$^*$  & 2.76\%$^*$        & 1.62  & 3.68  & 3.62\%       & 1.88  & 4.34  & 4.41\% \\
            &ST-Mamba      & 1.30$^*$  & 2.89  & 2.94\%       & 1.61$^*$  & 3.32$^*$  & 3.56\%$^*$       & 1.88  & 4.38  & 4.40\% \\
		\cmidrule(r){2-11}
		  &\textbf{STformer} & \underline{1.14} & \underline{2.32} & \underline{2.34}\% & \underline{1.37}  & \underline{2.92} & \underline{2.94}\%  & \underline{1.60}  & \underline{3.64}  & \underline{3.61}\% \\
            \cmidrule(r){2-11}
            &\textbf{NSTformer} & \textbf{1.14} & \textbf{2.31} & \textbf{2.33\%} & \textbf{1.36} & \textbf{2.91} & \textbf{2.92\%}  & \textbf{1.60} & \textbf{3.64} & \textbf{3.59\%} \\
			\midrule
		  \bottomrule
		\end{tabular}
	  \end{table*}

\subsection{Datasets}

\textbf{METR-LA} and \textbf{PEMS-BAY} are the two most commonly used datasets in traffic forecasting \cite{li2018diffusion}, we select these and give a brief introduction of the two datasets in \textbf{Table 2}.

\subsection{Experimental Setup}
In our experiments, METR-LA and PEMS-BAY datasets are partitioned into training, validation, and test sets with a ratio of 7:1:2.

Following \cite{liu2023spatio}, in our models, the embedding dimension $d_f$ is 24, while $d_a$ is set to 80. The number of Tranformer layers in STformer and its invariant in NSTformer are 3, all models with 4 heads. Both input and forecasting length are configured to be 1 hour, namely 12 time-steps. As for the landmarks, \cite{xiong2021nystromformer} states that 64 landmarks is sufficient to yield a robust approximation. In order to cluster all the nodes at each of the 12 time-steps, we set the number of landmarks to 72, namely, cluster the nodes to 6 neighborhoods. For cluster algorithm, we use Agglomerative Clustering from scikit-learn \cite{sklearn.AgglomerativeClustering}. 

We select Adam as optimizer, set learning rate to 0.001 and weight decay to 0.0003, train 30 epochs to optimize masked mae loss. We use the mostly used three metrics, MAE (Mean Absolute Error), RMSE (Root Mean Squared Error), and MAPE (Mean Absolute Percentage Error) to evaluating forecasting performance.

All experiments were conducted on a server with Ubuntu 18.04.1 and NVIDIA A100 GPU with 320 GB memory in total. We use Python 3.9.13, Pytorch 1.13.0 and BasicTS \cite{shao2023exploringprogressmultivariatetime} platform to run our models.

\begin{table}[htpb]
\renewcommand\arraystretch{0.98}
		\centering
		\setlength{\abovecaptionskip}{0.cm}
  \caption{Intro of Datasets}
  \label{tab:freq}
  \begin{tabular}{ccccc}
    \toprule
    \midrule
    \textbf{Datasets}&\textbf{Nodes}&\textbf{Steps}&\textbf{Freq}&\textbf{Range}\\
    \midrule
    \midrule
    METR-LA&207&34,272&5 mins&03-06/2012\\
    PEMS-BAY&325&52,116&5 mins&01-05/2017\\
    \midrule
  \bottomrule
\end{tabular}
\end{table}

\subsection{Baselines}

We selected abundant and diverse baselines, HA (Historical Average), HI \cite{cui2021historical}, SVR \cite{smola2004tutorial} and VAR \cite{lu2016integrating} are traditional forecasting methods. FC-LSTM \cite{sutskever2014sequence} is a typical deep learning method. DCRNN \cite{li2018diffusion}, AGCRN \cite{bai2020adaptive}, GTS \cite{shang2021discrete}, GWNet \cite{ijcai2019p264}, MTGNN \cite{wu2020connecting}, STGCN \cite{yu2018spatio} and STSGCN \cite{song2020spatial} are typical representative within the Spatial-Temporal Graph framework. Though with attention mechanism, ASTGCN \cite{guo2019attention}, GMAN \cite{zheng2020gman}, STID \cite{shao2022spatial}, STNorm \cite{deng2021st}, PDFormer \cite{pdformer} are still within the framework. STAEformer \cite{liu2023spatio} used vanilla Transformer with proposed spatio-temporal adaptive embedding earned excellent performance in traffic forecasting task, but captured the spatial-temporal representation separately. Actually, the comparision between STAEformer and STformer can be regarded as an ablation study. We have noticed that ST-Mamba \cite{shao2024stmambaspatialtemporalselectivestate} introduced Mamba \cite{gu2024mambalineartimesequencemodeling} to traffic forecasting with a linear model complexity, we select it to give a pivotal comparison between linear attention and Mamba in spatial-temporal forecasting task.

\subsection{Results Analysis}

\textbf{Table 1} shows our results. If STformer and NSTformer exceed all other models, we marked the results of STformer and NSTformer with underline and bold format respectively. In all cases, bold results are the best while results with asterisk represents the best model except STformer and NSTformer. 

As it shows, NSTformer and STformer achieve best performance and second best performance respectively on almost all metrics. At the first glance, it could be astonishing that NSTformer can be a bit better than STformer. In a few cases, \cite{xiong2021nystromformer} also discovered that Nystr{\" o}m attention is even slightly better than self-attention in NLP tasks. Based on the discovery, we propose an open problem here: \textbf{Does approximate attention have any additional positive effect such as regularization ?}

STAEformer and ST-Mamba are only inferior to NSTformer and STformer. Our work with with STAEformer reveal the power of pure attention in spatial-temporal forecasting. Due to learning the spatial and temporal relationships separately and asynchronously, our models exceed it with a big advantage. Though ST-Mamba also has linear complexity, it lags far behind our models.

\subsection{Model Complexity}

\begin{table}[htpb]
\renewcommand\arraystretch{0.98}
		\centering
		\setlength{\abovecaptionskip}{0.cm}
  \caption{Model Complexity on METR-LA}
  \label{tab:freq}
  \begin{tabular}{ccrc}
    \toprule
    \midrule
    \textbf{Models}&\textbf{Asymp}&\textbf{Params}&\textbf{Layers}\\
    \midrule
    \midrule
    STAEformer&$O(N^2+T^2)$&3,004,264&6\\
    STformer&$O(N^2T^2)$&743,388&3\\
    NSTformer&$O(NT)$&742,020&3\\
    \midrule
  \bottomrule
\end{tabular}
\end{table}

We present the complexity of STformer and NSTformer in \textbf{Table 3}, with the comparison to STAEformer, the parameters are recorded on METR-LA. STformer and NSTformer can capture spatial-temporal relationship between nodes with fewer layers thanks to the fully-connected setting, which yields fewer model parameters, indicating their afford cost further.

\section{Conclusion}

We investigated only using attention mechanism in spatial-temporal forecasting tasks to address the problems of existing works. Our STformer earn state-of-the-art performance with a large advantage. Given the quadratic complexity of Transformer and based on our insight to the instinct of spatial-temporal data, we propose NSTformer with linear complexity by adapting Nystr{\" o}mformer to overcome the obstacle, which slightly exceed STformer  surprisingly. Thanks to the leading performance and economical cost of NSTformer, we hope our work can offer insight to spatial-temporal forecasting. For future research, one can extend our method to more other spatial-temporal tasks. Another interesting direction is to try other linear attention or Mamba further. As for the cases where approximate attention beat standard self-attention slightly, we get a hypothesis after speculating the discovery : Approximate attention could have additional positive effect such as regularization. We propose the theoretically and practically meaningful open problem to machine learning community.

\section{Acknowledgements}

We are grateful to Le Zhang for his instructive insight. We would like to thank Zezhi Shao and Zheng Dong for their tirelessly help in experiments. We also thank Tsz Chiu Kwok who corrected a misunderstanding we had about spatial-temporal relationships captured by Transformer.

\bibliography{aaai25}

\begin{thebibliography}{73}
\providecommand{\natexlab}[1]{#1}

\bibitem[{Bai et~al.(2020)Bai, Yao, Li, Wang, and Wang}]{bai2020adaptive}
Bai, L.; Yao, L.; Li, C.; Wang, X.; and Wang, C. 2020.
\newblock Adaptive graph convolutional recurrent network for traffic forecasting.
\newblock \emph{Advances in neural information processing systems}, 33: 17804--17815.

\bibitem[{Beltagy, Peters, and Cohan(2020)}]{beltagy2020longformer}
Beltagy, I.; Peters, M.~E.; and Cohan, A. 2020.
\newblock Longformer: The long-document transformer.
\newblock \emph{arXiv preprint arXiv:2004.05150}.

\bibitem[{Cai et~al.(2020)Cai, Janowicz, Mai, Yan, and Zhu}]{cai2020traffic}
Cai, L.; Janowicz, K.; Mai, G.; Yan, B.; and Zhu, R. 2020.
\newblock Traffic transformer: Capturing the continuity and periodicity of time series for traffic forecasting.
\newblock \emph{Transactions in GIS}, 24(3): 736--755.

\bibitem[{Chen et~al.(2020)Chen, Lin, Li, Li, Zhou, and Sun}]{chen2020measuring}
Chen, D.; Lin, Y.; Li, W.; Li, P.; Zhou, J.; and Sun, X. 2020.
\newblock Measuring and relieving the over-smoothing problem for graph neural networks from the topological view.
\newblock In \emph{Proceedings of the AAAI conference on artificial intelligence}, volume~34, 3438--3445.

\bibitem[{Chen, O’Bray, and Borgwardt(2022)}]{chen2022structure}
Chen, D.; O’Bray, L.; and Borgwardt, K. 2022.
\newblock Structure-aware transformer for graph representation learning.
\newblock In \emph{International Conference on Machine Learning}, 3469--3489. PMLR.

\bibitem[{Chen et~al.(2021{\natexlab{a}})Chen, Lu, Rajeswaran, Lee, Grover, Laskin, Abbeel, Srinivas, and Mordatch}]{chen2021decision}
Chen, L.; Lu, K.; Rajeswaran, A.; Lee, K.; Grover, A.; Laskin, M.; Abbeel, P.; Srinivas, A.; and Mordatch, I. 2021{\natexlab{a}}.
\newblock Decision transformer: Reinforcement learning via sequence modeling.
\newblock \emph{Advances in neural information processing systems}, 34: 15084--15097.

\bibitem[{Chen et~al.(2021{\natexlab{b}})Chen, Zeng, Ji, and Yang}]{chen2021skyformer}
Chen, Y.; Zeng, Q.; Ji, H.; and Yang, Y. 2021{\natexlab{b}}.
\newblock Skyformer: Remodel self-attention with gaussian kernel and nystr$\backslash$" om method.
\newblock \emph{Advances in Neural Information Processing Systems}, 34: 2122--2135.

\bibitem[{Choromanski et~al.(2020)Choromanski, Likhosherstov, Dohan, Song, Gane, Sarlos, Hawkins, Davis, Mohiuddin, Kaiser et~al.}]{choromanski2020rethinking}
Choromanski, K.~M.; Likhosherstov, V.; Dohan, D.; Song, X.; Gane, A.; Sarlos, T.; Hawkins, P.; Davis, J.~Q.; Mohiuddin, A.; Kaiser, L.; et~al. 2020.
\newblock Rethinking Attention with Performers.
\newblock In \emph{International Conference on Learning Representations}.

\bibitem[{Cui, Xie, and Zheng(2021)}]{cui2021historical}
Cui, Y.; Xie, J.; and Zheng, K. 2021.
\newblock Historical inertia: A neglected but powerful baseline for long sequence time-series forecasting.
\newblock In \emph{Proceedings of the 30th ACM international conference on information \& knowledge management}, 2965--2969.

\bibitem[{Deng et~al.(2021)Deng, Chen, Jiang, Song, and Tsang}]{deng2021st}
Deng, J.; Chen, X.; Jiang, R.; Song, X.; and Tsang, I.~W. 2021.
\newblock St-norm: Spatial and temporal normalization for multi-variate time series forecasting.
\newblock In \emph{Proceedings of the 27th ACM SIGKDD conference on knowledge discovery \& data mining}, 269--278.

\bibitem[{Deshpande et~al.(2006)Deshpande, Rademacher, Vempala, and Wang}]{deshpande2006matrix}
Deshpande, A.; Rademacher, L.; Vempala, S.~S.; and Wang, G. 2006.
\newblock Matrix approximation and projective clustering via volume sampling.
\newblock \emph{Theory of Computing}, 2(1): 225--247.

\bibitem[{Devlin et~al.(2018)Devlin, Chang, Lee, and Toutanova}]{devlin2018bert}
Devlin, J.; Chang, M.-W.; Lee, K.; and Toutanova, K. 2018.
\newblock Bert: Pre-training of deep bidirectional transformers for language understanding.
\newblock \emph{arXiv preprint arXiv:1810.04805}.

\bibitem[{Dosovitskiy et~al.(2020)Dosovitskiy, Beyer, Kolesnikov, Weissenborn, Zhai, Unterthiner, Dehghani, Minderer, Heigold, Gelly et~al.}]{dosovitskiy2020image}
Dosovitskiy, A.; Beyer, L.; Kolesnikov, A.; Weissenborn, D.; Zhai, X.; Unterthiner, T.; Dehghani, M.; Minderer, M.; Heigold, G.; Gelly, S.; et~al. 2020.
\newblock An image is worth 16x16 words: Transformers for image recognition at scale.
\newblock \emph{arXiv preprint arXiv:2010.11929}.

\bibitem[{Farahat, Ghodsi, and Kamel(2011)}]{farahat2011novel}
Farahat, A.; Ghodsi, A.; and Kamel, M. 2011.
\newblock A novel greedy algorithm for Nystr{\"o}m approximation.
\newblock In \emph{Proceedings of the Fourteenth International Conference on Artificial Intelligence and Statistics}, 269--277. JMLR Workshop and Conference Proceedings.

\bibitem[{Feng et~al.(2022)Feng, Chen, Li, Sarkar, and Zhang}]{feng2022powerful}
Feng, J.; Chen, Y.; Li, F.; Sarkar, A.; and Zhang, M. 2022.
\newblock How powerful are k-hop message passing graph neural networks.
\newblock \emph{Advances in Neural Information Processing Systems}, 35: 4776--4790.

\bibitem[{Frieze, Kannan, and Vempala(2004)}]{frieze2004fast}
Frieze, A.; Kannan, R.; and Vempala, S. 2004.
\newblock Fast Monte-Carlo algorithms for finding low-rank approximations.
\newblock \emph{Journal of the ACM (JACM)}, 51(6): 1025--1041.

\bibitem[{Gittens and Mahoney(2016)}]{gittens2016revisiting}
Gittens, A.; and Mahoney, M.~W. 2016.
\newblock Revisiting the Nystr{\"o}m method for improved large-scale machine learning.
\newblock \emph{The Journal of Machine Learning Research}, 17(1): 3977--4041.

\bibitem[{Gu and Dao(2024)}]{gu2024mambalineartimesequencemodeling}
Gu, A.; and Dao, T. 2024.
\newblock Mamba: Linear-Time Sequence Modeling with Selective State Spaces.
\newblock arXiv:2312.00752.

\bibitem[{Guo et~al.(2019)Guo, Lin, Feng, Song, and Wan}]{guo2019attention}
Guo, S.; Lin, Y.; Feng, N.; Song, C.; and Wan, H. 2019.
\newblock Attention based spatial-temporal graph convolutional networks for traffic flow forecasting.
\newblock In \emph{Proceedings of the AAAI conference on artificial intelligence}, volume~33, 922--929.

\bibitem[{Han et~al.(2021)Han, Du, Sun, Fu, Lv, and Xiong}]{han2021dynamic}
Han, L.; Du, B.; Sun, L.; Fu, Y.; Lv, Y.; and Xiong, H. 2021.
\newblock Dynamic and multi-faceted spatio-temporal deep learning for traffic speed forecasting.
\newblock In \emph{Proceedings of the 27th ACM SIGKDD conference on knowledge discovery \& data mining}, 547--555.

\bibitem[{He et~al.(2016)He, Zhang, Ren, and Sun}]{he2016deep}
He, K.; Zhang, X.; Ren, S.; and Sun, J. 2016.
\newblock Deep residual learning for image recognition.
\newblock In \emph{Proceedings of the IEEE conference on computer vision and pattern recognition}, 770--778.

\bibitem[{Jiang et~al.(2023)Jiang, Han, Zhao, and Wang}]{pdformer}
Jiang, J.; Han, C.; Zhao, W.~X.; and Wang, J. 2023.
\newblock PDFormer: Propagation Delay-aware Dynamic Long-range Transformer for Traffic Flow Prediction.
\newblock In \emph{{AAAI}}. {AAAI} Press.

\bibitem[{Katharopoulos et~al.(2020)Katharopoulos, Vyas, Pappas, and Fleuret}]{katharopoulos2020transformers}
Katharopoulos, A.; Vyas, A.; Pappas, N.; and Fleuret, F. 2020.
\newblock Transformers are rnns: Fast autoregressive transformers with linear attention.
\newblock In \emph{International conference on machine learning}, 5156--5165. PMLR.

\bibitem[{Keles, Wijewardena, and Hegde(2023)}]{keles2023computational}
Keles, F.~D.; Wijewardena, P.~M.; and Hegde, C. 2023.
\newblock On the computational complexity of self-attention.
\newblock In \emph{International Conference on Algorithmic Learning Theory}, 597--619. PMLR.

\bibitem[{Kidger and Lyons(2020)}]{kidger2020universal}
Kidger, P.; and Lyons, T. 2020.
\newblock Universal approximation with deep narrow networks.
\newblock In \emph{Conference on learning theory}, 2306--2327. PMLR.

\bibitem[{Kim et~al.(2022)Kim, Nguyen, Min, Cho, Lee, Lee, and Hong}]{kim2022pure}
Kim, J.; Nguyen, D.; Min, S.; Cho, S.; Lee, M.; Lee, H.; and Hong, S. 2022.
\newblock Pure transformers are powerful graph learners.
\newblock \emph{Advances in Neural Information Processing Systems}, 35: 14582--14595.

\bibitem[{Kipf and Welling(2016)}]{kipf2016semi}
Kipf, T.~N.; and Welling, M. 2016.
\newblock Semi-Supervised Classification with Graph Convolutional Networks.
\newblock In \emph{International Conference on Learning Representations}.

\bibitem[{Kitaev, Kaiser, and Levskaya(2019)}]{kitaev2019reformer}
Kitaev, N.; Kaiser, L.; and Levskaya, A. 2019.
\newblock Reformer: The Efficient Transformer.
\newblock In \emph{International Conference on Learning Representations}.

\bibitem[{Kumar, Mohri, and Talwalkar(2009)}]{kumar2009ensemble}
Kumar, S.; Mohri, M.; and Talwalkar, A. 2009.
\newblock Ensemble nystrom method.
\newblock \emph{Advances in Neural Information Processing Systems}, 22.

\bibitem[{Kumar, Mohri, and Talwalkar(2012)}]{kumar2012sampling}
Kumar, S.; Mohri, M.; and Talwalkar, A. 2012.
\newblock Sampling methods for the Nystr{\"o}m method.
\newblock \emph{The Journal of Machine Learning Research}, 13(1): 981--1006.

\bibitem[{Lee et~al.(2019)Lee, Lee, Kim, Kosiorek, Choi, and Teh}]{pmlr-v97-lee19d}
Lee, J.; Lee, Y.; Kim, J.; Kosiorek, A.; Choi, S.; and Teh, Y.~W. 2019.
\newblock Set Transformer: A Framework for Attention-based Permutation-Invariant Neural Networks.
\newblock In Chaudhuri, K.; and Salakhutdinov, R., eds., \emph{Proceedings of the 36th International Conference on Machine Learning}, volume~97 of \emph{Proceedings of Machine Learning Research}, 3744--3753. PMLR.

\bibitem[{Li, Kwok, and L{\"u}(2010)}]{li2010making}
Li, M.; Kwok, J. T.-Y.; and L{\"u}, B. 2010.
\newblock Making large-scale Nystr{\"o}m approximation possible.
\newblock In \emph{Proceedings of the 27th International Conference on Machine Learning, ICML 2010}, 631.

\bibitem[{Li et~al.(2019)Li, Jin, Xuan, Zhou, Chen, Wang, and Yan}]{li2019enhancing}
Li, S.; Jin, X.; Xuan, Y.; Zhou, X.; Chen, W.; Wang, Y.-X.; and Yan, X. 2019.
\newblock Enhancing the locality and breaking the memory bottleneck of transformer on time series forecasting.
\newblock \emph{Advances in neural information processing systems}, 32.

\bibitem[{Li et~al.(2018)Li, Yu, Shahabi, and Liu}]{li2018diffusion}
Li, Y.; Yu, R.; Shahabi, C.; and Liu, Y. 2018.
\newblock Diffusion Convolutional Recurrent Neural Network: Data-Driven Traffic Forecasting.
\newblock In \emph{International Conference on Learning Representations}.

\bibitem[{Likhosherstov et~al.(2021)Likhosherstov, Choromanski, Davis, Song, and Weller}]{likhosherstov2021sub}
Likhosherstov, V.; Choromanski, K.~M.; Davis, J.~Q.; Song, X.; and Weller, A. 2021.
\newblock Sub-linear memory: How to make performers slim.
\newblock \emph{Advances in Neural Information Processing Systems}, 34: 6707--6719.

\bibitem[{Lin et~al.(2023)Lin, Li, Li, Bai, Zhao, and Zhang}]{lin2023dynamic}
Lin, J.; Li, Z.; Li, Z.; Bai, L.; Zhao, R.; and Zhang, C. 2023.
\newblock Dynamic causal graph convolutional network for traffic prediction.
\newblock \emph{arXiv preprint arXiv:2306.07019}.

\bibitem[{Liu et~al.(2023)Liu, Dong, Jiang, Deng, Deng, Chen, and Song}]{liu2023spatio}
Liu, H.; Dong, Z.; Jiang, R.; Deng, J.; Deng, J.; Chen, Q.; and Song, X. 2023.
\newblock Spatio-temporal adaptive embedding makes vanilla transformer sota for traffic forecasting.
\newblock In \emph{Proceedings of the 32nd ACM International Conference on Information and Knowledge Management}, 4125--4129.

\bibitem[{Liu et~al.(2021)Liu, Lin, Cao, Hu, Wei, Zhang, Lin, and Guo}]{liu2021swin}
Liu, Z.; Lin, Y.; Cao, Y.; Hu, H.; Wei, Y.; Zhang, Z.; Lin, S.; and Guo, B. 2021.
\newblock Swin transformer: Hierarchical vision transformer using shifted windows.
\newblock In \emph{Proceedings of the IEEE/CVF international conference on computer vision}, 10012--10022.

\bibitem[{Livni, Shalev-Shwartz, and Shamir(2014)}]{livni2014computational}
Livni, R.; Shalev-Shwartz, S.; and Shamir, O. 2014.
\newblock On the computational efficiency of training neural networks.
\newblock \emph{Advances in neural information processing systems}, 27.

\bibitem[{Lu et~al.(2021)Lu, Yao, Zhang, Zhu, Xu, Gao, Xu, Xiang, and Zhang}]{lu2021soft}
Lu, J.; Yao, J.; Zhang, J.; Zhu, X.; Xu, H.; Gao, W.; Xu, C.; Xiang, T.; and Zhang, L. 2021.
\newblock Soft: Softmax-free transformer with linear complexity.
\newblock \emph{Advances in Neural Information Processing Systems}, 34: 21297--21309.

\bibitem[{Lu and Lu(2020)}]{lu2020universal}
Lu, Y.; and Lu, J. 2020.
\newblock A universal approximation theorem of deep neural networks for expressing probability distributions.
\newblock \emph{Advances in neural information processing systems}, 33: 3094--3105.

\bibitem[{Lu et~al.(2016)Lu, Zhou, Wu, Jiang, and Cui}]{lu2016integrating}
Lu, Z.; Zhou, C.; Wu, J.; Jiang, H.; and Cui, S. 2016.
\newblock Integrating Granger Causality and Vector Auto-Regression for Traffic Prediction of Large-Scale WLANs.
\newblock \emph{KSII Transactions on Internet \& Information Systems}, 10(1).

\bibitem[{Peng et~al.(2021)Peng, Pappas, Yogatama, Schwartz, Smith, and Kong}]{peng2021random}
Peng, H.; Pappas, N.; Yogatama, D.; Schwartz, R.; Smith, N.; and Kong, L. 2021.
\newblock Random Feature Attention.
\newblock In \emph{International Conference on Learning Representations (ICLR 2021)}.

\bibitem[{Razavi et~al.(2014)Razavi, Kerayechian, Gachpazan, and Shateyi}]{razavi2014new}
Razavi, M.~K.; Kerayechian, A.; Gachpazan, M.; and Shateyi, S. 2014.
\newblock A New Iterative Method for Finding Approximate Inverses of Complex Matrices.
\newblock In \emph{Abstract and Applied Analysis}, volume 2014, 1--7. Hindawi.

\bibitem[{Roy et~al.(2021)Roy, Saffar, Vaswani, and Grangier}]{roy2021efficient}
Roy, A.; Saffar, M.; Vaswani, A.; and Grangier, D. 2021.
\newblock Efficient content-based sparse attention with routing transformers.
\newblock \emph{Transactions of the Association for Computational Linguistics}, 9: 53--68.

\bibitem[{scikit-learn developers(2024)}]{sklearn.AgglomerativeClustering}
scikit-learn developers. 2024.
\newblock Agglomerative Clustering.
\newblock \url{https://scikit-learn.org/stable/modules/generated/sklearn.cluster.AgglomerativeClustering.html}.
\newblock Accessed: 2024-08-16.

\bibitem[{Shang and Chen(2021)}]{shang2021discrete}
Shang, C.; and Chen, J. 2021.
\newblock Discrete Graph Structure Learning for Forecasting Multiple Time Series.
\newblock In \emph{Proceedings of International Conference on Learning Representations}.

\bibitem[{Shao et~al.(2024)Shao, Bell, Wang, Geers, Xi, and Gao}]{shao2024stmambaspatialtemporalselectivestate}
Shao, Z.; Bell, M. G.~H.; Wang, Z.; Geers, D.~G.; Xi, H.; and Gao, J. 2024.
\newblock ST-Mamba: Spatial-Temporal Selective State Space Model for Traffic Flow Prediction.
\newblock arXiv:2404.13257.

\bibitem[{Shao et~al.(2023)Shao, Wang, Xu, Wei, Yu, Zhang, Yao, Jin, Cao, Cong, Jensen, and Cheng}]{shao2023exploringprogressmultivariatetime}
Shao, Z.; Wang, F.; Xu, Y.; Wei, W.; Yu, C.; Zhang, Z.; Yao, D.; Jin, G.; Cao, X.; Cong, G.; Jensen, C.~S.; and Cheng, X. 2023.
\newblock Exploring Progress in Multivariate Time Series Forecasting: Comprehensive Benchmarking and Heterogeneity Analysis.
\newblock arXiv:2310.06119.

\bibitem[{Shao et~al.(2022{\natexlab{a}})Shao, Zhang, Wang, Wei, and Xu}]{shao2022spatial}
Shao, Z.; Zhang, Z.; Wang, F.; Wei, W.; and Xu, Y. 2022{\natexlab{a}}.
\newblock Spatial-temporal identity: A simple yet effective baseline for multivariate time series forecasting.
\newblock In \emph{Proceedings of the 31st ACM International Conference on Information \& Knowledge Management}, 4454--4458.

\bibitem[{Shao et~al.(2022{\natexlab{b}})Shao, Zhang, Wei, Wang, Xu, Cao, and Jensen}]{shao2022decoupled}
Shao, Z.; Zhang, Z.; Wei, W.; Wang, F.; Xu, Y.; Cao, X.; and Jensen, C.~S. 2022{\natexlab{b}}.
\newblock Decoupled dynamic spatial-temporal graph neural network for traffic forecasting.
\newblock \emph{arXiv preprint arXiv:2206.09112}.

\bibitem[{Shen et~al.(2018)Shen, Wang, Wang, Min, Su, Zhang, Li, Henao, and Carin}]{shen-etal-2018-baseline}
Shen, D.; Wang, G.; Wang, W.; Min, M.~R.; Su, Q.; Zhang, Y.; Li, C.; Henao, R.; and Carin, L. 2018.
\newblock Baseline Needs More Love: On Simple Word-Embedding-Based Models and Associated Pooling Mechanisms.
\newblock In Gurevych, I.; and Miyao, Y., eds., \emph{Proceedings of the 56th Annual Meeting of the Association for Computational Linguistics (Volume 1: Long Papers)}, 440--450. Melbourne, Australia: Association for Computational Linguistics.

\bibitem[{Si, Hsieh, and Dhillon(2016)}]{si2016computationally}
Si, S.; Hsieh, C.-J.; and Dhillon, I. 2016.
\newblock Computationally efficient Nystr{\"o}m approximation using fast transforms.
\newblock In \emph{International conference on machine learning}, 2655--2663. PMLR.

\bibitem[{Smola and Sch{\"o}lkopf(2004)}]{smola2004tutorial}
Smola, A.~J.; and Sch{\"o}lkopf, B. 2004.
\newblock A tutorial on support vector regression.
\newblock \emph{Statistics and computing}, 14: 199--222.

\bibitem[{Song et~al.(2020)Song, Lin, Guo, and Wan}]{song2020spatial}
Song, C.; Lin, Y.; Guo, S.; and Wan, H. 2020.
\newblock Spatial-temporal synchronous graph convolutional networks: A new framework for spatial-temporal network data forecasting.
\newblock In \emph{Proceedings of the AAAI conference on artificial intelligence}, volume~34, 914--921.

\bibitem[{Sutskever, Vinyals, and Le(2014)}]{sutskever2014sequence}
Sutskever, I.; Vinyals, O.; and Le, Q.~V. 2014.
\newblock Sequence to sequence learning with neural networks.
\newblock \emph{Advances in neural information processing systems}, 27.

\bibitem[{Valiant(1984)}]{valiant1984theory}
Valiant, L.~G. 1984.
\newblock A theory of the learnable.
\newblock \emph{Communications of the ACM}, 27(11): 1134--1142.

\bibitem[{Vaswani et~al.(2017)Vaswani, Shazeer, Parmar, Uszkoreit, Jones, Gomez, Kaiser, and Polosukhin}]{NIPS2017_3f5ee243}
Vaswani, A.; Shazeer, N.; Parmar, N.; Uszkoreit, J.; Jones, L.; Gomez, A.~N.; Kaiser, L.~u.; and Polosukhin, I. 2017.
\newblock Attention is All you Need.
\newblock In Guyon, I.; Luxburg, U.~V.; Bengio, S.; Wallach, H.; Fergus, R.; Vishwanathan, S.; and Garnett, R., eds., \emph{Advances in Neural Information Processing Systems}, volume~30. Curran Associates, Inc.

\bibitem[{Vyas, Katharopoulos, and Fleuret(2020)}]{NEURIPS2020_f6a8dd1c}
Vyas, A.; Katharopoulos, A.; and Fleuret, F. 2020.
\newblock Fast Transformers with Clustered Attention.
\newblock In Larochelle, H.; Ranzato, M.; Hadsell, R.; Balcan, M.; and Lin, H., eds., \emph{Advances in Neural Information Processing Systems}, volume~33, 21665--21674. Curran Associates, Inc.

\bibitem[{Wang et~al.(2023)Wang, Jiang, Jiang, Han, and Zhao}]{wang2023towards}
Wang, J.; Jiang, J.; Jiang, W.; Han, C.; and Zhao, W.~X. 2023.
\newblock Towards Efficient and Comprehensive Urban Spatial-Temporal Prediction: A Unified Library and Performance Benchmark.
\newblock \emph{arXiv preprint arXiv:2304.14343}.

\bibitem[{Wang et~al.(2020)Wang, Li, Khabsa, Fang, and Ma}]{wang2020linformer}
Wang, S.; Li, B.~Z.; Khabsa, M.; Fang, H.; and Ma, H. 2020.
\newblock Linformer: Self-attention with linear complexity.
\newblock \emph{arXiv preprint arXiv:2006.04768}.

\bibitem[{Wu et~al.(2020{\natexlab{a}})Wu, Xiao, Ding, Zhao, Wei, and Huang}]{wu2020adversarial}
Wu, S.; Xiao, X.; Ding, Q.; Zhao, P.; Wei, Y.; and Huang, J. 2020{\natexlab{a}}.
\newblock Adversarial sparse transformer for time series forecasting.
\newblock \emph{Advances in neural information processing systems}, 33: 17105--17115.

\bibitem[{Wu et~al.(2020{\natexlab{b}})Wu, Pan, Long, Jiang, Chang, and Zhang}]{wu2020connecting}
Wu, Z.; Pan, S.; Long, G.; Jiang, J.; Chang, X.; and Zhang, C. 2020{\natexlab{b}}.
\newblock Connecting the dots: Multivariate time series forecasting with graph neural networks.
\newblock In \emph{Proceedings of the 26th ACM SIGKDD international conference on knowledge discovery \& data mining}, 753--763.

\bibitem[{Wu et~al.(2019)Wu, Pan, Long, Jiang, and Zhang}]{ijcai2019p264}
Wu, Z.; Pan, S.; Long, G.; Jiang, J.; and Zhang, C. 2019.
\newblock Graph WaveNet for Deep Spatial-Temporal Graph Modeling.
\newblock In \emph{Proceedings of the Twenty-Eighth International Joint Conference on Artificial Intelligence, {IJCAI-19}}, 1907--1913. International Joint Conferences on Artificial Intelligence Organization.

\bibitem[{Xiong et~al.(2021)Xiong, Zeng, Chakraborty, Tan, Fung, Li, and Singh}]{xiong2021nystromformer}
Xiong, Y.; Zeng, Z.; Chakraborty, R.; Tan, M.; Fung, G.; Li, Y.; and Singh, V. 2021.
\newblock Nystr{\"o}mformer: A nystr{\"o}m-based algorithm for approximating self-attention.
\newblock In \emph{Proceedings of the AAAI Conference on Artificial Intelligence}, volume~35, 14138--14148.

\bibitem[{Xu et~al.(2020)Xu, Dai, Liu, Gao, Lin, Qi, and Xiong}]{xu2020spatial}
Xu, M.; Dai, W.; Liu, C.; Gao, X.; Lin, W.; Qi, G.-J.; and Xiong, H. 2020.
\newblock Spatial-temporal transformer networks for traffic flow forecasting.
\newblock \emph{arXiv preprint arXiv:2001.02908}.

\bibitem[{Yu, Yin, and Zhu(2018)}]{yu2018spatio}
Yu, B.; Yin, H.; and Zhu, Z. 2018.
\newblock Spatio-temporal graph convolutional networks: a deep learning framework for traffic forecasting.
\newblock In \emph{Proceedings of the 27th International Joint Conference on Artificial Intelligence}, 3634--3640.

\bibitem[{Yun et~al.(2019)Yun, Jeong, Kim, Kang, and Kim}]{yun2019graph}
Yun, S.; Jeong, M.; Kim, R.; Kang, J.; and Kim, H.~J. 2019.
\newblock Graph transformer networks.
\newblock \emph{Advances in neural information processing systems}, 32.

\bibitem[{Zaheer et~al.(2020)Zaheer, Guruganesh, Dubey, Ainslie, Alberti, Ontanon, Pham, Ravula, Wang, Yang et~al.}]{zaheer2020big}
Zaheer, M.; Guruganesh, G.; Dubey, K.~A.; Ainslie, J.; Alberti, C.; Ontanon, S.; Pham, P.; Ravula, A.; Wang, Q.; Yang, L.; et~al. 2020.
\newblock Big bird: Transformers for longer sequences.
\newblock \emph{Advances in neural information processing systems}, 33: 17283--17297.

\bibitem[{Zhang, Tsang, and Kwok(2008)}]{10.1145/1390156.1390311}
Zhang, K.; Tsang, I.~W.; and Kwok, J.~T. 2008.
\newblock Improved Nystr\"{o}m low-rank approximation and error analysis.
\newblock In \emph{Proceedings of the 25th International Conference on Machine Learning}, ICML '08, 1232–1239. New York, NY, USA: Association for Computing Machinery.
\newblock ISBN 9781605582054.

\bibitem[{Zhao et~al.(2017)Zhao, Chen, Wu, Chen, and Liu}]{zhao2017lstm}
Zhao, Z.; Chen, W.; Wu, X.; Chen, P.~C.; and Liu, J. 2017.
\newblock LSTM network: a deep learning approach for short-term traffic forecast.
\newblock \emph{IET Intelligent Transport Systems}, 11(2): 68--75.

\bibitem[{Zheng et~al.(2020)Zheng, Fan, Wang, and Qi}]{zheng2020gman}
Zheng, C.; Fan, X.; Wang, C.; and Qi, J. 2020.
\newblock Gman: A graph multi-attention network for traffic prediction.
\newblock In \emph{Proceedings of the AAAI conference on artificial intelligence}, volume~34, 1234--1241.

\bibitem[{Zhou et~al.(2021)Zhou, Zhang, Peng, Zhang, Li, Xiong, and Zhang}]{zhou2021informer}
Zhou, H.; Zhang, S.; Peng, J.; Zhang, S.; Li, J.; Xiong, H.; and Zhang, W. 2021.
\newblock Informer: Beyond efficient transformer for long sequence time-series forecasting.
\newblock In \emph{Proceedings of the AAAI conference on artificial intelligence}, volume~35, 11106--11115.

\end{thebibliography}

\end{document}